%% file: main.tex
\documentclass[conference]{IEEEtran}

\ifCLASSINFOpdf
\else
   \usepackage[dvips]{graphicx}
\fi
\usepackage{url}
\usepackage{amsmath}
\usepackage{amsthm}
\usepackage{amssymb}
\usepackage{upgreek}
\usepackage{cite}
\usepackage{algorithm}
\usepackage{algorithmic}
\usepackage{bm}

\usepackage{graphicx}
\usepackage[utf8]{inputenc}

\begin{document}
\title{SMDP-Based Dynamic Batching for Efficient Inference on GPU-Based Platforms}  
\author{
    \IEEEauthorblockN{Yaodan Xu, Jingzhou Sun, Sheng Zhou, Zhisheng Niu}
    \IEEEauthorblockA{Beijing National Research Center for Information Science and Technology}
    \IEEEauthorblockA{Department of Electronic Engineering, Tsinghua University, Beijing 100084, P.R. China}
    
    \IEEEauthorblockA{\{xyd21, sunjz18\}@mails.tsinghua.edu.cn, \{sheng.zhou, niuzhs\}@tsinghua.edu.cn}
}
\maketitle


\input{Data/abstract_v2}

\input{Data/introduction}

\input{Data/model2}

\input{Data/control}

\input{Data/solution}
\input{Data/results}
\input{Data/conclusion_short}
\input{Data/acknowlegement}

\bibliographystyle{ieeetr}  
\bibliography{references}

\end{document}

%% file: Data/abstract_v2.tex
\begin{abstract}
In up-to-date machine learning (ML) applications on cloud or edge computing platforms, batching is an important technique for providing efficient and economical services at scale.
In particular, parallel computing resources on the platforms, such as graphics processing units (GPUs), have higher computational and energy efficiency with larger batch sizes.
However, larger batch sizes may also result in longer response time, and thus it requires a judicious design.
This paper aims to provide a dynamic batching policy that strikes a balance between efficiency and latency.
The GPU-based inference service is modeled as a batch service queue with batch-size dependent processing time.
Then, the design of dynamic batching is a continuous-time average-cost problem, and is formulated as a semi-Markov decision process (SMDP) with the objective of minimizing the weighted sum of average response time and average power consumption.
The optimal policy is acquired by solving an associated discrete-time Markov decision process (MDP) problem with finite state approximation and ``discretization".
By introducing an abstract cost to reflect the impact of ``tail" states, the space complexity and the time complexity of the procedure can decrease by 63.5\% and 98\%, respectively.
Our results show that the optimal policies potentially possess a control limit structure.
Numerical results also show that SMDP-based batching policies can adapt to different traffic intensities and outperform other benchmark policies.
Furthermore, the proposed solution has notable flexibility in balancing power consumption and latency.
\end{abstract}

%% file: Data/introduction.tex
\section{Introduction}

The last decade has witnessed the rapid growth of machine learning (ML), and graphics processing units (GPUs) play a prominent role in accelerating the training and inference of neural networks due to their advantage in parallel computing\cite{oh2004gpu}. 
The wide deployment of ML on various devices sparks a growing demand for ML-as-a-Service (MLaaS) platforms such as Google Cloud Prediction\cite{GoogleCloud}, where trained models are published in the cloud to provide
inference (prediction) services for massive end-users.

\begin{figure}
    \centering    \includegraphics[width=0.64\linewidth]{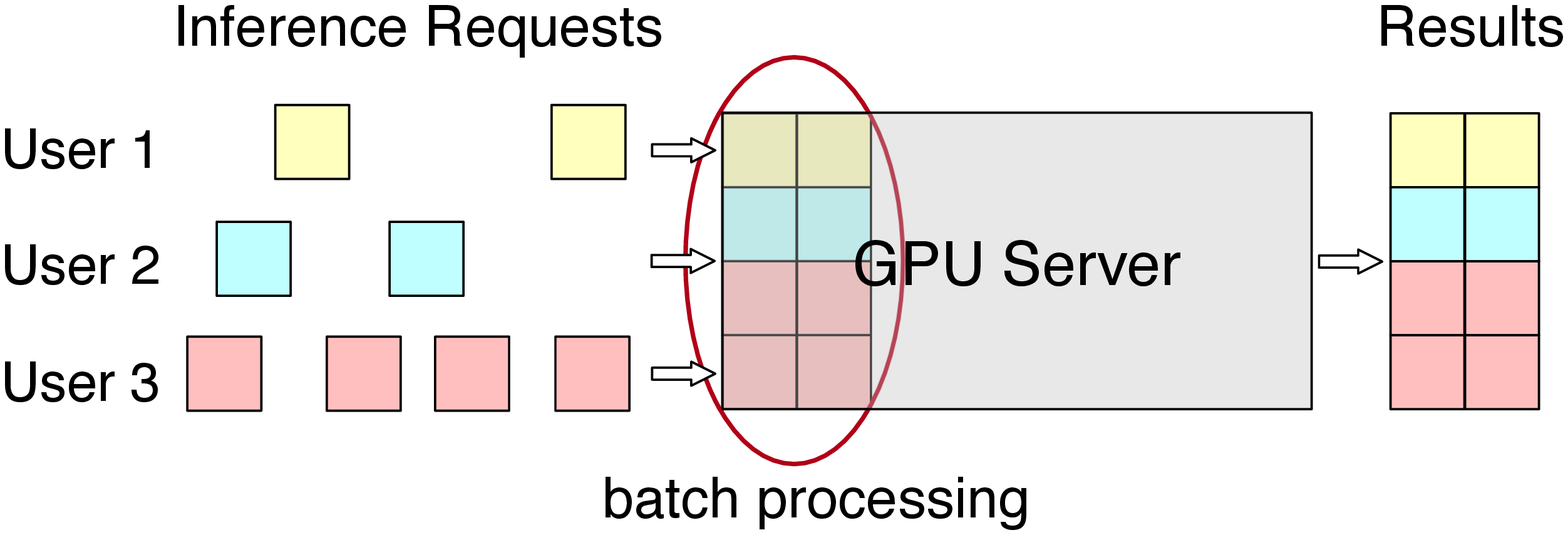}
    \caption{Batching of the inference requests from different users on a GPU-based platform.}
    \label{fig:batching}
\end{figure}

An important factor that affects both cost and performance of ML inference serving is batch processing, or \emph{batching}\cite{zhang2019mark,crankshaw2017clipper}.
For GPUs, batching brings about a significant increase in both computational efficiency and energy efficiency\cite{NVIDIA},
i.e., the average processing time (energy consumption) per task decreases with the batch size. 
As depicted in Fig.~\ref{fig:batching}, the GPU-based computing platform can bundle the same type of inference requests from possibly different users, and serve them concurrently using the same ML application, for example in the scenario of Function as a Service (FaaS)\cite{castro2019rise}.

Despite the well-known benefits, there are two main challenges in batching:
1) Although batching improves the throughput of GPU, the inference response time is likely to increase because of waiting to form a batch, as well as longer processing time of serving more requests at once\cite{ali2020batch,crankshaw2017clipper}.
2) Static configured batching does not fit realistic scenarios\cite{choi2021lazy}, where it shows poor responsiveness at low load and low throughput at high load\cite{cui20202}. 
To tackle these problems, a dynamic batching scheme is required, to judiciously adjust the batch size to balance the performance and cost.


Recently, several dynamic batching schemes have been proposed for machine learning inference systems\cite{crankshaw2017clipper,serf,zhang2019mark,cui20202,choi2021lazy,ali2020batch,yao2022eais}.
Meanwhile, only a small amount of researches have made progresses in the  performance analysis. 
SERF\cite{serf} models the inference serving as a M/D/c queue, but unfortunately batching is not explicitly considered in the model.
Another work BATCH\cite{ali2020batch} models the request arrival as a Poisson process or Markov-modulated Poisson
process with two phases (MMPP(2))\cite{lu2013approach} and gets the number of requests that arrive before timeout.
However, the cases where requests arrive during processing time are missing from the analysis.
The author of \cite{inoue2021queueing} gives a closed-form queuing analysis of dynamic batching under the work-conserving policy, which is not optimal since it eliminates the potential of forming larger batches.
  

In fact, dynamic batching decision is an optimal control problem in batch service queues.
Although there are abundant researches concerning this topic in operational research\cite{deb1973optimal,papadaki2002exploiting,fowler2021survey}, they do not cope with our problem because of their essential assumption that the processing time of a batch is \emph{independent} of its
batch size.
In \cite{inoue2022load}, the author models the load-balancing problem in multiple batch service queues with \emph{size-dependent} processing time as a Markov decision process (MDP) and claims it as an open problem.
Actually, optimal batching in one such batch service queue, as the simplest sub-problem of \cite{inoue2022load},  still remains unsolved. 

To the best of our knowledge, our work is the first to formulate and optimally solve the dynamic batching of inference requests as a semi-Markov decision process (SMDP).
The considered objective is a weighted sum of average response time and average power consumption.
To solve the problem, we provide a procedure composed of finite state approximation, model ``discretization" and relative value iteration.
The demanding problem of state space explosion is tackled by a novel introduction of an abstract cost, which reflects the impact of costs in ``tail" states. 
From numerical results, we observe that the optimal policies potentially possess a control limit structure, which could inspire the simplification of representation and computation in future researches.  
Comparisons with other benchmarks demonstrate that: 1) The SMDP-based policies always achieve the best objective in different settings of parameters.
2) When having the same average response time, the SMDP-based policies never consume more energy than any other benchmark policy, and vice versa.  
Moreover, the proposed solution can adapt to different traffic intensities, and it can also flexibly balance the response time and power consumption.





%% file: Data/model2.tex
\section{System Model}\label{sec:system2}
We consider the scenario of a single GPU-based server in the continuous-time setting.
Batching is executed on the same type of inference requests from possibly different end-users, and
it cannot be interrupted until the current batch is processed.
The system is modeled as a single service queue with the arrival of inference requests following a Poisson process with intensity $\lambda$.
The requests waiting to be processed are stored in a buffer which is assumed to be infinitely large.
The total number of requests in the buffer as well as being processed at time $t \geq 0$ is denoted by $s(t) \in \{0,1,2,\dots\}$.

\begin{figure}
    \centering    \includegraphics[width=0.65\linewidth]{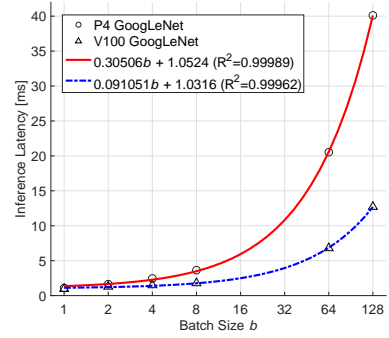}
    \caption{Inference latency for batch processing GoogLeNet\cite{szegedy2015going} on TESLA P4 and TESLA V100 respectively. 
    The batch size is plotted in $\log_2$ coordinate.}
    \label{fig:latency}
\end{figure}

Many researchers have conducted experiments to profile the computation latency of GPU when batch processing ML inference tasks\cite{ali2020batch,inoue2021queueing,shi2022multi,serf,hanhirova2018latency,crankshaw2017clipper,choi2021lazy,yao2022eais,cui20202}.
It has been examined in \cite{ali2020batch} that the coefficient of variation (CV) of the batch processing time for image recognition is near CV=0, i.e. \emph{deterministic}.
This result is reasonable since the calculations in ML inference tasks are predefined.
Meanwhile, profiling results also exhibit that the inference latency grows \emph{linearly} with the batch size\cite{inoue2021queueing,cui20202,yao2022eais}. 
As shown in Fig.~\ref{fig:latency}, linear regression on the statistics from NIVIDIA\cite{NVIDIA} also validates the conclusion.

\begin{figure}
    \centering
    \includegraphics[width=0.65\linewidth]{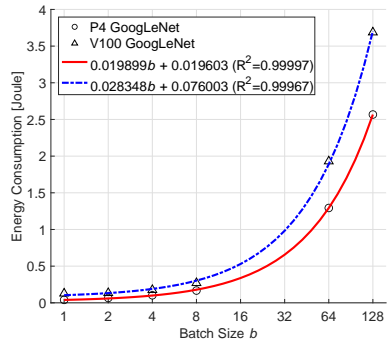}
    \caption{Energy consumption for batch processing GoogLeNet\cite{szegedy2015going} on TESLA P4 and TESLA V100 respectively.
    The batch size is plotted in $\log_2$ coordinate.
    }
    \label{fig:energy}
\end{figure}


Let $b \in \mathcal{B} \triangleq \{1,2,\ldots,B_{\max}\}$ denote the batch size, where $B_{\max}$ is the maximum batch size allowed by the system.
We assume that the batch processing time is a deterministic value $\tau^{[b]}$ and is linear with the batch size $b$, given by
\begin{equation}
\tau^{[b]}=\alpha b+\tau_0,
\end{equation}
for some $\alpha>0$ and $\tau_0 \geq 0$.
Define ${\mu}^{[b]}=b/\tau^{[b]}$ as the mean throughput (the number of inference requests processed per unit time) for a batch size $b$.
It is easy to verify that the mean throughput $\mu^{[b]}$ is non-decreasing with the batch size $b$, and the maximum throughput is acquired at the maximum batch size $B_{\max}$.
Let $\rho=\lambda/\mu^{[B_{\max}]}$ denote the ratio of the arrival rate over the maximum throughput.
We assume that $\rho$ satisfies $\rho \in [0,1)$, which is a necessary condition for
the system stability.

The energy consumption of processing a batch of $b$ requests is denoted by ${\zeta}^{[b]}$, which can be calculated by the product of the average board power and the batch processing time.
Also based on the statistics from \cite{NVIDIA}, we perform linear regression as shown in Fig.~\ref{fig:energy} and
make the following assumption that the energy consumption ${\zeta}^{[b]}$ is linear with the batch size $b$, which is given by
\begin{equation}
{\zeta}^{[b]}=\beta b+{\zeta}_0,
\end{equation}
for some $\beta>0$ and ${\zeta}_0 \geq 0$.
The energy efficiency $\eta^{[b]}=b / {\zeta}^{[b]}$ is defined as the average number of requests served with one unit of energy consumption.
It is clear that the energy efficiency $\eta^{[b]}$ is non-decreasing with the batch size $b$.

As a result, given the GPU and the inference model, the parameters $B_{\max}, \alpha, \tau_0, \beta, {\zeta}_0$ can be profiled and therefore determined.




There are two main factors to be considered in the objective: One is the request response time, or latency, which includes both waiting and processing time, as the performance metric. 
The other is power consumption on the server, as the running cost metric.
Our objective is to minimize the average weighted sum of these two metrics.

The serving process consists of sequential service rounds.
Define $t_i\; (t_i \geq 0, \;i=1,2,\ldots)$ as the start time of the $i$th service round and $b(t_i) \in \mathcal{B}$ as the batch size in the $i$th service round.
Let $N(t) \in \mathbb{N}$ denote the total number of service rounds until time $t \geq 0$.
The objective is then expressed as
\begin{equation}
        \min \; \limsup\limits_{T\rightarrow\infty} \frac{1}{T}
        \Bigg\{
        w_1 \frac{1}{\lambda} \int_{0}^{T} s(t) \, \mathrm{d}t
        +
        w_2\sum_{i=1}^{N(T)} {\zeta}^{[b(t_i)]}
        \Bigg\}
        ,
    \label{obj1}
\end{equation}
where $w_1 \geq 0$ and $w_2 \geq 0$ are the weights.
Note that here the average request response time is equivalently transformed to the average queue length by Little's Law\cite{little1961proof}.

%% file: Data/control.tex
\section{SMDP Formulation}\label{sec:formulation}
It is natural to formulate (\ref{obj1}) as a continuous-time system and utilize the embedded Markov chain method, or semi-Markov decision process (SMDP).
The continuous-time SMDP formulation reduces the system state from three elements to one element, compared to the discrete-time MDP formulation\cite{inoue2022load}. 
As a result, it can relieve the curse of dimensionality when applying iteration-based algorithms.

In SMDP, we only consider the states at decision points, when either the server completes a batch of service, or a request arrives while the server is idle.
The decision points divide the timeline into sequential decision epochs.
Let $\{s_m,m=0,1,\ldots\}$ denote the state process, where the state is the number of requests in the system, taking values from the state space $\mathcal{S} \triangleq \{0,1,2,\ldots\}$.
At each epoch $m$, the server takes an action $a_m$ from the
action space $\mathcal{A} \triangleq \{0,1,2,\ldots,B_{\max}\}$.
The action $a_m$ is the size of batch to be processed, where $a_m=0$ means to keep idle in the $m$th epoch.
Let $\mathcal{A}_s \subseteq \mathcal{A}$ be the set of feasible actions for a given state $s$.
Note that the number of requests to be batched should be no more than the available requests, which means $\mathcal{A}_s \triangleq \{0,1,2,\ldots,\min\{s,B_{\max}\}\}$.

The state transition is associated with the action.
Let $m(j|s,a)$ denote the probability that the semi-Markov decision process occupies state $j$ at the next decision epoch when action $a$ is chosen at the state $s$.
Let $p_k^{[b]}$ denote the probability that $k$ requests arrive during the period of processing a batch of $b \in \mathcal{B}$ requests.
Since the arrival of requests follows a Poisson process, we have
\begin{equation}
    p_k^{[b]} 
    = \frac{e^{-\lambda {\tau}^{[b]}}(\lambda {\tau}^{[b]})^k}{k !}
    , \quad k=0,1, \ldots
\end{equation}
Then the transition probability for $\forall j,s \in \mathcal{S}, \forall a \in \mathcal{A}_s$ is expressed as
\begin{equation}
m(j|s,a)=
\begin{cases}
p_{j-s+a}^{[a]}& \text{ $ j \geq s-a,$ $a \in \mathcal{A}_s,$ $a \neq 0$ } \\
1& \text{ $ j = s+1,$ $a=0$ }
\\
0& \text{ otherwise }
\end{cases}.
\end{equation}

Let a random variable $\gamma_m$ denote the sojourn time between the
$(m-1)$th and the $m$th epoch.
Let $\delta(\cdot)$ denote the impulse function.
The probability density function (PDF) of $\gamma$ in state $s$ under action $a$ is denoted by $f_{s,a}(x)\; (x \geq 0)$, given by
\begin{equation}
    f_{s,a}(x) =
    \begin{cases}
   \delta\left(x-{\tau}^{[a]}\right)& \quad  a \in \mathcal{A}_s, a \neq 0
    \\
    \lambda e^{-\lambda x}&
    \quad  a = 0
    \end{cases},
    \; \forall s \in \mathcal{S}.
\end{equation}

Define $y(s, a)=\mathbb{E}[\gamma|s,a]$ as the expected sojourn time until the next decision epoch, given by
\begin{equation}
    y(s,a) =
    \begin{cases}
   {\tau}^{[a]}& \quad  a \in \mathcal{A}_s, a \neq 0
    \\
    1/\lambda&
    \quad  a = 0
    \end{cases},
    \; \forall s \in \mathcal{S}.
    \label{eqy}
\end{equation}

Costs are charged for serving the requests as well as holding them.
The cost of serving a batch of $x$ requests is denoted by $u(x)$, and the cost of holding $x$ requests in the system \emph{per unit time} is denoted by $v(x)$.
Let $c(s, a)$ denote the expected cost until the next decision epoch when action $a \in \mathcal{A}_s$ is taken in state $s$. We have $c(s,0)=u(0)+v(s)y(s,0)$, and for $a \neq 0$,
\begin{equation}
c(s,a)=
  u(a)+
\int_0^{\infty}\int_0^{x}
f_{s,a}(x)\sum_{k=0}^{\infty} v(s+k) \frac{e^{-\lambda t}(\lambda t)^k}{k !} \mathrm{d} t
\mathrm{d} x.
\end{equation}

The cost functions corresponding to the objective in (\ref{obj1}) are $u(x)=w_2 {\zeta}^{[x]} \; (x>0)$, $u(0)=0$  and $v(x)=\frac{w_1}{\lambda} x$.
This leads to a detailed description of $c(s,a)$ as
\begin{equation}
\begin{aligned}
    c(s,a)=& \;
    u(a)+
    \int_0^{{\tau}^{[a]}} \sum_{k=0}^{\infty} v(s+k) \frac{e^{-\lambda t}(\lambda t)^k}{k !} \mathrm{d} t
    \\
    =& \;
    w_2{\zeta}^{[a]} +
    \int_0^{{\tau}^{[a]}} \frac{w_1}{\lambda}(s+\lambda t)
    \mathrm{d} t
    \\
    =& \;
    w_2{\zeta}^{[a]} +
    {w_1}[\frac{s}{\lambda}{\tau}^{[a]}+\frac{1}{2}{({\tau}^{[a]})}^2]
    , \quad a \in \mathcal{A}_s, a \neq 0,
    \\
    c(s,0) =& \; w_1\frac{s}{{\lambda}^2}.
\end{aligned}
\end{equation}

A decision rule $\bm{d}^{(m)} \in {\mathbb{R}}^{\left|\mathcal{A}\right|}$ is a vector of probabilities assigned to each action at epoch $m$, and it is \emph{deterministic} if one action is taken with probability 1.
A policy $\bm{\pi}=\{\bm{d}^{(0)},\bm{d}^{(1)},\bm{d}^{(2)},\ldots \}$ is a sequence of decision rules, and it is called \emph{stationary} if $\bm{d}^{(m)}=\bm{d}, \forall m \in \mathbb{N}$.
Let $n_t$ denote the number of decisions up to time $t$.
Our goal is to find a policy that minimizes the long term average expected
cost $g^{\bm{\pi}}(s_0)$, given that the system occupies state $s_0 \in \mathcal{S}$ at $t=0$, which is
\begin{equation}
   \min \limits_{\bm{\pi}}
   g^{\bm{\pi}}(s_0)=
   \limsup\limits_{T\rightarrow\infty} \frac{1}{T}
        {\mathbb{E}}^{\bm{\pi}}_s
        \Bigg\{
         \int_{0}^{T} v\left(s(t)\right) \, \mathrm{d}t
        +
        \sum_{i=0}^{n_T-1} u(a_i)
        \Bigg\}
        .
\label{objective2}
\end{equation}

Since $\mathcal{S}$ is countable and the model can be easily verified to be unichain, according to \cite{puterman2014markov}, the optimality equations are
\begin{equation}
h(s)=\min _{a \in A_s}\Bigg\{c(s, a)-g y(s, a)+\sum_{j \in S} m(j | s, a) h(j)\Bigg\},
\label{eq7}
\end{equation}
for $\forall s \in \mathcal{S}$, where $h(s)$ denotes the relative value of being in state $s$ and $g$ denotes the average cost per unit time.
By Theorem 11.4.4 in \cite{puterman2014markov}, the constant and functions $(g,h)$ satisfying (\ref{eq7}) are exactly the optimal average cost per unit time and the corresponding relative value functions. 

The considered model is an infinite state SMDP with non-negative, unbounded costs and finite action sets.
Sennott has proved in \cite{sennott1989average} that an average expected optimal \emph{stationary deterministic} policy exists in such a model under several conditions that firmly hold in our problem. 
The details are omitted here due to the page limits.



%% file: Data/solution.tex
\section{Procedure for Solving Infinite State SMDP}
\subsection{Finite State Approximation}
\label{subsec:approximate}

The SMDP problem has infinite states in $\mathcal{S}=\{0,1,2,\ldots\}$, and is impractical to be solved by numerical methods.
Hence, we truncate the infinite state space to a \emph{finite} state space $\mathcal{S}'=\{0,1,\ldots,s_{\max},S_{o}\}$, which replaces the states larger than $s_{\max}$ by an ``overflow" state $S_{o}$.
The dimension of the finite state space is $|\mathcal{S}'|=s_{\max}+2$, where $s_{\max}$ needs to be at least no less than $B_{\max}$.
The rationale of the state space truncation is that the tail probability, defined as the probability of being in the ``tail" states $\mathcal{\hat{S}}=\{s_{\max}+1,s_{\max}+2,\ldots\}$, decreases with $s_{\max}$.
When $s_{\max}$ is large enough, the ``tail" states are negligible.
Detailed analysis of convergence and error bounds of finite state approximation can be found in \cite{thomas1985finite}.

In the truncated model, the action space $\mathcal{A}$, the sojourn time distribution $f_{s,a}$, and the expected sojourn time $y(s,a)$ are the same as before, while the feasible action space at state $S_o$ is $\mathcal{A}_{S_o}=\{0,1,\ldots,B_{\max}\} 
\equiv \mathcal{A}$ since $s_{\max} \geq B_{\max}$.
Original transitions to the ``tail" states $\mathcal{\hat{S}}$ are aggregated to $S_{o}$, and
we assume the number of requests at $S_{o}$ is  $s_{\max}$. 
The adapted transition probability $m'(j|s,a)$ for $\forall j,s \in \mathcal{S'}, \forall a \in \mathcal{A}_s$ is
\begin{equation}
\begin{aligned}
&m'(j|S_o,a) =\\
&\begin{cases}
p_{j-s_{\max}+a}^{[a]}&  j \geq s_{\max}-a, j \neq S_{o},a \neq 0\\
1-\sum\limits_{i=0}^{a}p_i^{[a]}
& j = S_{o},a \neq 0 \\
1& j=S_{o},a=0
\\
0& \text{ otherwise }
\end{cases},
\\
&m'(j|s,a) \; (s \neq S_o)=\\
&\begin{cases}
p_{j-s+a}^{[a]}& j \geq s-a, j \neq S_{o},a \neq 0 \\
1-\sum\limits_{i=0}^{s_{\max}-s+a}p_i^{[a]}
&
j = S_{o},a \neq 0 \\
1& j = s+1,s<s_{\max},a=0
\\
1& j=S_{o},s = s_{\max},a=0
\\
0& \text{ otherwise }
\end{cases}.
\end{aligned}
\label{eqm}
\end{equation}

The unbounded holding cost induced by the infinite states in the primal problem is also erased by the truncation.
Therefore, we introduce an abstract cost $c_oy(s,a) \; (c_o \geq 0)$ to the ``overflow" state $S_o$, working as an estimation of the difference between the expected holding cost at ``tail" states and the holding cost at $s_{\max}$.
The adapted cost $c'(s,a)$ is
\begin{equation}
\begin{aligned}
c'(s,a)= 
\begin{cases}
    c(s_{\max},a)+c_oy(s,a) \quad s = S_o \\
    c(s,a) \quad \quad \quad \quad \; s \neq S_o, s \in \mathcal{S'}
\end{cases},
\end{aligned}
\forall a \in \mathcal{A}_s.
\label{eq13}
\end{equation}
Since $\rho \in [0,1)$, the optimal policy must stabilize the system.
The abstract cost can be also interpreted as an overflow punishment, which pushes the optimal policy away from causing overflow.
Note that the abstract cost $c_oy(s,a)$ in (\ref{eq13}) is rarely mentioned in the literature, without which the problem can be solved as well, but with a larger satisfactory $s_{\max}$ and higher computational complexity in iteration algorithms (which will be discussed in section~\ref{subsec:complexity}).

We establish a criterion to assess the approximation regarding the difference in average cost: Given a stationary deterministic policy as a function $\pi:\mathcal{S'} \rightarrow \mathcal{A}$, the corresponding state transition matrix is $P_{{\pi}} \in \mathbb{R}^{{|\mathcal{S'}|}\times{|\mathcal{S'}|}}$.
Suppose that the Markov chain with $P_{\pi}$ has a unique stationary distribution $\bm{\mu}=(\mu_0,\mu_1,\ldots,\mu_{S_o})$.
Then the average cost per unit time is
\begin{equation}
    g^{\pi}= \frac{\sum_{s \in \mathcal{S'}} \mu_s \cdot c'(s,\pi(s))}{\sum_{s \in \mathcal{S'}} \mu_s \cdot y(s,\pi(s)) }.
    \label{eq14}
\end{equation}
Let $\Delta^{\pi}$ be the average cost contributed by $S_o$ per unit time:
\begin{equation}
    \Delta^{\pi}= \frac{\mu_{S_o} \cdot c'(S_o,\pi(S_o))}{\sum_{s \in \mathcal{S'}} \mu_s \cdot y(s,\pi(s)) }.
\label{eq15}
\end{equation}
Given a predefined constant $\delta>0$, if $\Delta^{\pi} < \delta$, we state that the approximation is acceptable with tolerance $\delta$.
Otherwise, the approximation is not acceptable with tolerance $\delta$ and a larger $s_{\max}$ should be selected.
 
The optimality equations of the finite state SMDP are
\begin{equation}
\begin{aligned}
h(s)=\min _{a \in A_s}\Bigg\{c'(s, a)-g y(s, a)+\sum_{j \in S'} m'(j | s, a) h(j)\Bigg\},
\end{aligned}
\label{optimality3}
\end{equation}
for $\forall s \in \mathcal{S'}$.
Denote $g^*$ as the optimal average expected cost.

\begin{figure*}
\centering
\includegraphics[width=0.70\linewidth]{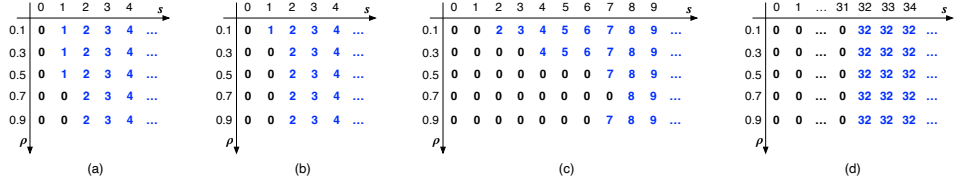}
\caption{The optimal policies for the average cost problem (\ref{objective2}) with energy and latency characteristics of GoogLeNet on TESLA P4.
The maximum batch size is chosen as $B_{\max}=32$.
The weights are
(a) $w_1 = 1, w_2 = 0$; (b) $w_1 = 1, w_2 = 0.1$; (c) $w_1 = 1, w_2 = 1$; (d) $w_1 = 1, w_2 = 500$.} 
\label{policy_view}
\end{figure*}

\subsection{Associated Discrete-Time MDP}
\label{subsec:dtmdp}
The finite state continuous-time SMDP is associated with a \emph{discrete-time} MDP through a ``discretization" transformation (See Chapter 11.4 in \cite{puterman2014markov}).
The state space $\mathcal{S'}$, the action space $\mathcal{A}$ and the feasible action space $\mathcal{A}_s$ for any $s \in \mathcal{S'}$ keep unchanged in the transformed model.
The transformed cost $\tilde{c}(s, a)$ and the transformed transition probability $\tilde{m}(j | s, a)$ for $\forall j,s \in \mathcal{S'}, \forall a \in \mathcal{A}_s$ are
\begin{equation}
\begin{aligned}
    \tilde{c}(s, a) \triangleq& \; c'(s, a) / y(s, a),\\
    \tilde{m}(j | s, a) \triangleq& 
    \begin{cases}
    \eta m'(j | s, a) / y(s, a)& j \neq s \\ 
    1+\eta[m'(s| s, a)-1] / y(s, a)& j=s
    \end{cases},
\end{aligned}
\end{equation}
where $\eta$ satisfies
\begin{equation}
    0<\eta<y(s, a)/(1-m'(s|s, a)),
\end{equation}
for all $a \in A_s$ and $s \in S'$ for which $m'(s|s, a)<1$. 

By (\ref{eqy}) and (\ref{eqm}), an appropriate $\eta$ should satisfy
\begin{equation}
  0<\eta<\min\Bigg\{\frac{1}{\lambda},\min_{a \in \mathcal{A}}\bigg\{
  \frac{{\tau}^{[a]}}{1-p_a^{[a]}}, \frac{{\tau}^{[a]}}{\sum\limits_{i=0}^{a}p_i^{[a]}}
  \bigg\}\Bigg\}.  
\end{equation}
And from experiments we find that the larger the $\eta$ is, the faster the value-based iteration algorithm converges.

For the discrete-time MDP, the optimality equations are
\begin{equation}
h(s)=\min _{a \in {\mathcal{A}}_s}\Bigg\{\tilde{c}(s,a)-g +\sum_{j \in S'} \tilde{m}(j|s, a) h(j)\Bigg\},
\forall s \in \mathcal{S'}.
\label{eq12}
\end{equation}
By Proposition 11.4.5 in \cite{puterman2014markov}, if $(\tilde{g},\tilde{h})$ satisfy the discrete-time optimality equations in (\ref{eq12}), then $(\tilde{g},\eta\tilde{h})$ satisfy (\ref{optimality3}) and $\tilde{g}=g^*$ is the optimal average cost in the continuous-time model.
And Theorem 11.4.6 in \cite{puterman2014markov} proves the existence of an average optimal stationary deterministic policy.





\subsection{Relative Value Iteration}
\label{subsec:rvi}
In this paper, we utilize the value-based iteration algorithm to solve the discrete-time MDP.
Specifically, for the average-cost MDP, the standard value iteration is numerically unstable, so we use relative value iteration (RVI)\cite{puterman2014markov} instead.

Let $J_{n}(s)$ be the value function at state $s \in \mathcal{S'}$ in the $n$th iteration, where $n \in \mathbb{N}$.
At the beginning of the algorithm, a state $s^* \in \mathcal{S'}$ is arbitrarily chosen.
The iterative formula of RVI is 
\begin{equation}
    J_{n+1} (s)=\min _{a \in A_s}\Bigg\{\tilde{c}(s, a)-J_{n} (s^*)+\sum_{j \in S'} \tilde{m}(j | s, a) J_{n}(j)\Bigg\}.
    \label{eq21}
\end{equation}
Note that in each iteration, the value function of $s^*$ is subtracted from the standard Bellman equation.
And the computed policy consists of $a^*(s)$ that minimizes the RHS of (\ref{eq21}).

In practice, we specify a maximum number of iterations, which is $\rm{iter}_{\max}$.
Another stopping criterion is that the span of the difference between iterations is smaller than a predefined constant $\epsilon>0$, which is expressed as

\begin{equation}
    \max \limits_{s \in \mathcal{S'}} \{J_{n+1} (s)-J_{n} (s)\} -
    \min \limits_{s \in \mathcal{S'}} \{J_{n+1} (s)-J_{n} (s)\}
    <
    \epsilon.
\end{equation}

Suppose the total number of iterations is $n$. 
The number of multiplications per iteration is $\sum \limits_{s \in \mathcal{S'}} |\mathcal{A}_s||\mathcal{S'}| \approx B_{\max}s_{\max}^2$.
Then, the time complexity is $\mathcal{O}(nB_{\max}s^2_{\max})$.
The space complexity is mainly influenced by $\tilde{c}(s, a)$ and $\tilde{m}(j | s, a)$.
Referring to (\ref{eqm}), the storage of $\tilde{m}(j | s, a)$ reduces to the storage of $p_i^{[a]}$.
Therefore, the space complexity is $\mathcal{O}(B_{\max}s_{\max})$.

As aforementioned, the optimal policy of the discrete-time MDP (in section~\ref{subsec:dtmdp}) is also optimal in its associated finite state SMDP (in section~\ref{subsec:approximate}).
Furthermore, it should be optimal in the original infinite state SMDP (in section~\ref{sec:formulation}) as well if the finite state approximation (in section~\ref{subsec:approximate}) is accurate enough.
Moreover, the computational complexity can be decreased if a smaller $s_{\max}$ is used in the approximation.

%% file: Data/results.tex
\section{Numerical Results}\label{sec:results}
\begin{figure}
\centering
\includegraphics[width=0.62\linewidth]{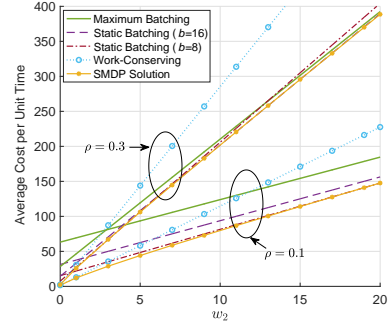}
\caption{The comparison of different policies on the average cost per unit time, where $w_1=1$ and $w_2$ varies between $0$ and $20$.} \label{comparison}
\end{figure}


In this section, we take the GoogLeNet inference on TESLA P4 as the model, where the batch processing latency is $\tau^{[b]}=0.3051b+1.052$ (ms) and the energy consumption is ${\zeta}^{[b]}=19.90 b+19.60$ (mJ).
The maximum batch size $B_{\max}$ is taken as 32 and the maximum service rate is $\mu^{[B_{\max}]}=2.96$ (requests/ms).
The average power consumption is measured by (mJ/ms), which is Watt (W).
We use $\rho \in [0,1)$ to represent the traffic intensity normalized by $\mu^{[B_{\max}]}$. 

\subsection{SMDP Solution}
Fig.~\ref{policy_view} demonstrates the policies obtained by solving the SMDP under different $\rho$ and $w_2$.
Each row in the chart is a stationary deterministic policy, where each element denotes the action taken at the state corresponding to the column. 
It can be concluded from the solutions that the system should not serve until the state exceeds a threshold, which is called a \emph{control limit}.
This is exactly the conclusion in batch service queues with i.i.d. batch processing time\cite{deb1973optimal}.
The control limit structure generally exists in SMDP solutions from the numerous results we have obtained.
From Fig.~\ref{policy_view}, we find that the control limit increases with $w_2$.
When $w_2$ is as large as 500, the control limits under different traffic intensities are all $B_{\max}$.  
This is reasonable because the importance of power consumption grows with $w_2$, and the energy is better saved with a larger batch size. 
Another observation is that when $w_1=1$ and $w_2=0$, i.e. the objective is only concerned with latency, the control limits are close to 1.
It means that in this parameter setting, the increased computational efficiency rarely compensates the additional latency introduced by a larger control limit.
\subsection{Performance Comparison}
\begin{table*}[h!]\tiny
\caption{Evaluation of approximations acceptable with tolerance $\delta=0.001$ under different $c_o$.}
    \centering
    \begin{tabular}{cccccc}
       \hline
       $c_o$  & $10000$ & $1000$ & $100$ & $10$ & $0$ \\
       \hline
       $\min s_{\max}$ & $89$ & $78$ & $\bm{70}$ & $161$ & $192$ \\
       $\text{Iterations}$ & $1847$ & $1635$ & $\bm{1483}$ & $10000$ & $10000$ \\
       $\text{Space Complexity}$ & $2848$ & $2496$ & $\bm{2240}$ & $5152$ & $6144$ \\
       $\text{Time Complexity}$ & $4.68 \times 10^{8}$ & $3.18 \times 10^{8}$ & $\bm{2.33 \times 10^{8}}$ & $8.29 \times 10^{9}$ & $1.18 \times 10^{10}$ \\
       $\Delta^{\pi}$ & $9.36 \times 10^{-4}$ & $9.78 \times 10^{-4}$ & $8.36 \times 10^{-4}$ & $6.14 \times 10^{-12}$ & $1.30 \times 10^{-14}$ \\
       $g^{\pi}$ & $66.1384$ & $66.1383$ & $66.1377$ & $66.1374$ & $66.1374$  \\
       \hline
    \end{tabular}
    \label{table}
\end{table*}

\begin{figure}
\centering
\includegraphics[width=0.65\linewidth]{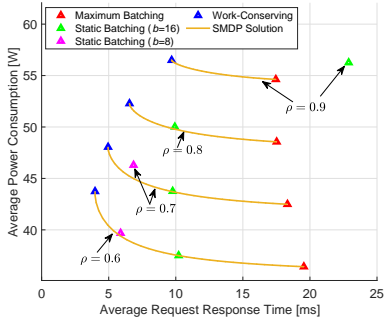}
\caption{The tradeoff between request response time and power consumption.} \label{tradeoff}
\end{figure}

\begin{figure}
\centering
\includegraphics[width=0.65\linewidth]{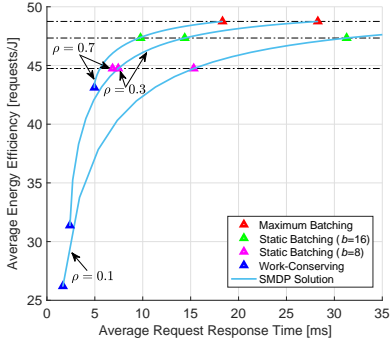}
\caption{The tradeoff between request response time and energy efficiency.} \label{tradeoff2}
\end{figure}

We compare the SMDP solutions with the work-conserving policy and static batching policies with batch size $b=8,16,32$.
In the work-conserving policy, the server processes the maximum feasible batch of current requests and never waits for a new-coming request unless there are no requests waiting.  
In static batching policies, the server always processes batches in a constant batch size, and waits for new-coming requests if there are not enough requests.
The static batching with $b=32$ is a special case called maximum batching policy in this setting where $B_{\max}=32$.
The objectives are computed using (\ref{eq14}).
It can be seen from Fig.~\ref{comparison} that the SMDP solution achieves the best among all policies under various parameter settings.
The work-conserving policy is far from optimal when $w_2>5$ and the maximum batching policy works badly when $\rho=0.1$.
The other two static batching policies can approach the optimal policy under certain parameter scales.

By going through different $w_1,w_2$, the optimal policies of the weighted-objective SMDP can balance latency and energy flexibly, where the pairs are formed into tradeoff curves as shown in Fig.~\ref{tradeoff} and Fig.~\ref{tradeoff2}.
The performance pairs of other policies are separate points on the figures since they do not change with the weights. 
The latency-energy pairs of the work-conserving policy and maximum batching policy are near the two end points of SMDP's tradeoff curve, which agrees with our observation in Fig.~\ref{policy_view} (a) and (d).
%
Several points of static batching policies are close to SMDP's tradeoff curve, meaning that these policies can well approximate the SMDP solutions under certain parameters, which agrees with the observation in Fig.~\ref{comparison}.
In other cases, the advantage of SMDP-based policies over static batching can be evidently observed.
For example, static batching with $b=8$ is above (below) the SMDP curve in Fig.~\ref{tradeoff} (Fig.~\ref{tradeoff2}) when $\rho=0.7$, which means that it consumes more energy (has lower energy efficiency) than a SMDP-based policy that has the same latency.
And it does not even stabilize the system when $\rho \geq 0.8$.
Similarly, static batching with $b=16$ has considerably longer latency when $\rho=0.9$ (see Fig.~\ref{tradeoff}).

\subsection{Approximation Accuracy and Complexity}
\label{subsec:complexity}
\begin{figure}
\centering
\includegraphics[width=0.65\linewidth]{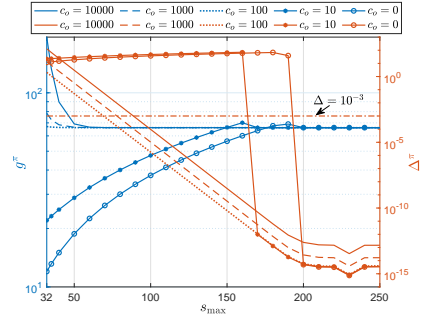}
\caption{The evolution of $g^{\pi}$ (the average cost per unit time) and $\Delta^{\pi}$ (the average cost contributed by $S_o$ per unit time) regarding $s_{\max}$ under different $c_o$, where $s_{\max}$ and $c_o$ are the parameters in finite state approximation.}
\label{convergence}
\end{figure}

We want to evaluate the accuracy and complexity of different finite space approximations.
As mentioned in section~\ref{subsec:approximate}, there are two parameters in the approximation: $s_{\max}$ and $c_o$, which control the dimension of the state space and the abstract cost, respectively.
Given the approximate model with certain $s_{\max}$ and $c_o$, a policy $\pi$ is computed following the procedure in section~\ref{subsec:dtmdp},~\ref{subsec:rvi}, and the average cost per unit time $g^{\pi}$ can be obtained.
The accuracy of approximation is evaluated by $\Delta^{\pi}$, which is the average cost contributed by $S_o$ per unit time.
The more accurate the approximation is, the less the $S_o$ (as the aggregation of the ``tail" states) should contribute, and thus the smaller the $\Delta^{\pi}$ is.
We conduct experiments in the setting of $\rho=0.9$ and $[w_1,w_2]=[1,1]$.
The stopping parameters in RVI are chosen as $\epsilon=0.01$ and $\rm{iter}_{\max}=10000$.

In Fig.~\ref{convergence}, we exhibit the evolution of $g^{\pi}$ and $\Delta^{\pi}$ with $s_{\max}$ from 32 to 250.
The $g^{\pi}$ with $c_o=10000, 1000, 100$ decreases and converges around $s_{\max}=35, 50, 70$, while $g^{\pi}$ with $c_o=10, 0$ increases and converges around $s_{\max}=170, 200$.
We can infer that the abstract cost with $c_o=10000, 1000, 100 \; (10,0)$ overestimates (underestimates) the impact of ``tail" states, leading to the $g^{\pi}$ mostly larger (smaller) than the convergence value.
From the orange curves, we see that $\Delta^{\pi}$ decreases with $s_{\max}$, and there exists a sharp drop for $c_o=10, 0$.
It is because the underestimated impact of ``tail" states with $c_o=10,0$ leads to an underestimated value for waiting ($a=0$) in the RHS of (\ref{eq21}).
And the computed policy is to always wait, until the $s_{\max}$ is large enough so that the cost of waiting is comparable with the cost of serving.
The $\Delta^{\pi}$ is near stable when $s_{\max}$ exceeds 200, and it increases with $c_o$ since the $\bm{\mu}$ and $y$ in (\ref{eq15}) are almost the same for all stable cases.
Although the stable $\Delta^{\pi}$ is no more than $10^{-13}$, we only need an approximation acceptable with tolerance $\delta$, and we choose $\delta=0.001$.
We list the minimum values of $s_{\max}$ that satisfy the approximation requirement in Table~\ref{table}.
The $\Delta^{\pi}$ and $g^{\pi}$, the number of RVI iterations, as well as the space and time complexity that correspond to the minimum $s_{\max}$ are also recorded.
It can be seen that all the $\Delta^{\pi}$ are less than 0.001, and the differences between $g^{\pi}$ are also less than 0.001.
The least required $s_{\max}$ is 70, which appears in the approximation with $c_o=100$.
Compared to the ordinary finite space approximation with $c_o=0$, the minimum $s_{\max}$ decreases from 192 to 70, the space complexity reduces by $63.5\%$, and the time complexity reduces by $98\%$ when $c_o=100$.
Furthermore, approximations with larger (smaller) $c_o$ show an increasing trend in complexity due to the growing overestimation (underestimation).
In practice, we need to choose an adequate $c_o$ to effectively reduce the complexity.

%% file: Data/conclusion_short.tex
\section{Conclusion}\label{sec:conclusion}
In this paper, we studied the problem of dynamic batching for machine learning inference on GPU-based platforms.
The problem is modeled as an infinite state semi-Markov decision process (SMDP) with the weighted sum of average response time and average power consumption as the objective. 
The SMDP is solved with finite state approximation, ``discretization" transformation and relative value iteration.
The computational complexity is largely reduced owning to the introduction of an abstract cost.
We take the GoogLeNet inference on TESLA P4 as an example and conduct extensive numerical experiments in various parameter settings. 
Numerical results have validated the superiority of the SMDP solution.
Compared to existing dynamic batching schemes, our proposed solution is theoretically derived, rather than simulation tryouts.
As a result, our scheme can be computed offline and release the system from the burden of extra complex modules.
The limitation is that our method only considers the average objective, instead of hard Service-Level
Objective (SLO) constraints, for which the proposed solution can function as a basic guideline and needs to be modified while running in real time.

%% file: Data/acknowlegement.tex
\section*{Acknowledgment}
The paper is sponsored in part by the National Key R$\&$D Program of China No. 2020YFB1806605, by the National Natural Science Foundation of China under Grants 62022049 and 62111530197, and by Hitachi.

%% file: main.bbl
\begin{thebibliography}{10}

\bibitem{oh2004gpu}
K.-S. Oh and K.~Jung, ``{GPU} implementation of neural networks,'' {\em Pattern
  Recognition}, vol.~37, no.~6, pp.~1311--1314, 2004.

\bibitem{GoogleCloud}
``Google cloud prediction {API} documentation.''
  \url{https://cloud.google.com/prediction/docs/}, 2017.
\newblock (accessed 19-Oct-2022).

\bibitem{zhang2019mark}
C.~Zhang, M.~Yu, W.~Wang, and F.~Yan, ``{MArk}: Exploiting cloud services for
  {Cost-Effective}, {SLO-Aware} machine learning inference serving,'' in {\em
  2019 USENIX Annual Technical Conference (USENIX ATC 19)}, pp.~1049--1062,
  2019.

\bibitem{crankshaw2017clipper}
D.~Crankshaw, X.~Wang, G.~Zhou, M.~J. Franklin, J.~E. Gonzalez, and I.~Stoica,
  ``Clipper: A {Low-Latency} online prediction serving system,'' in {\em 14th
  USENIX Symposium on Networked Systems Design and Implementation (NSDI 17)},
  pp.~613--627, 2017.

\bibitem{NVIDIA}
``{NVIDIA} {AI} inference platform technical overview.''
  \url{https://www.nvidia.com/en-us/data-center/resources/inference-technical-overview/},
  2018.
\newblock (accessed 23-Nov-2019).

\bibitem{castro2019rise}
P.~Castro, V.~Ishakian, V.~Muthusamy, and A.~Slominski, ``The rise of
  serverless computing,'' {\em Communications of the ACM}, vol.~62, no.~12,
  pp.~44--54, 2019.

\bibitem{ali2020batch}
A.~Ali, R.~Pinciroli, F.~Yan, and E.~Smirni, ``{BATCH}: Machine learning
  inference serving on serverless platforms with adaptive batching,'' in {\em
  SC20: International Conference for High Performance Computing, Networking,
  Storage and Analysis}, pp.~1--15, IEEE, 2020.

\bibitem{choi2021lazy}
Y.~Choi, Y.~Kim, and M.~Rhu, ``Lazy batching: An {SLA}-aware batching system
  for cloud machine learning inference,'' in {\em 2021 IEEE International
  Symposium on High-Performance Computer Architecture (HPCA)}, pp.~493--506,
  IEEE, 2021.

\bibitem{cui20202}
W.~Cui, Q.~Chen, H.~Zhao, M.~Wei, X.~Tang, and M.~Guo, ``E$^{\textrm{2}}$ bird:
  Enhanced elastic batch for improving responsiveness and throughput of deep
  learning services,'' {\em IEEE Trans. Parallel Distrib. Syst.}, vol.~32,
  no.~6, pp.~1307--1321, 2020.

\bibitem{serf}
F.~Yan, O.~Ruwase, Y.~He, and E.~Smirni, ``{SERF}: Efficient scheduling for
  fast deep neural network serving via judicious parallelism,'' in {\em SC16:
  Proceedings of the International Conference for High Performance Computing,
  Networking, Storage and Analysis}, pp.~300--311, 2016.

\bibitem{yao2022eais}
C.~Yao, W.~Liu, W.~Tang, and S.~Hu, ``{EAIS}: {Energy}-aware adaptive
  scheduling for {CNN} inference on high-performance {GPUs},'' {\em Future
  Generation Computer Systems}, vol.~130, pp.~253--268, 2022.

\bibitem{lu2013approach}
X.~Lu, J.~Yin, H.~Chen, and X.~Zhao, ``An approach for bursty and self-similar
  workload generation,'' in {\em International Conference on Web Information
  Systems Engineering}, pp.~347--360, Springer, 2013.

\bibitem{inoue2021queueing}
Y.~Inoue, ``Queueing analysis of {GPU}-based inference servers with dynamic
  batching: A closed-form characterization,'' {\em Performance Evaluation},
  vol.~147, p.~102183, 2021.

\bibitem{deb1973optimal}
R.~K. Deb and R.~F. Serfozo, ``Optimal control of batch service queues,'' {\em
  Advances in Applied Probability}, vol.~5, no.~2, pp.~340--361, 1973.

\bibitem{papadaki2002exploiting}
K.~P. Papadaki and W.~B. Powell, ``Exploiting structure in adaptive dynamic
  programming algorithms for a stochastic batch service problem,'' {\em
  European Journal of Operational Research}, vol.~142, no.~1, pp.~108--127,
  2002.

\bibitem{fowler2021survey}
J.~W. Fowler and L.~Mönch, ``A survey of scheduling with parallel batch
  (p-batch) processing,'' {\em European Journal of Operational Research},
  vol.~298, pp.~1--24, Apr. 2022.

\bibitem{inoue2022load}
Y.~Inoue, ``A load-balancing problem for distributed bulk-service queues with
  size-dependent batch processing times,'' {\em Queueing Systems}, vol.~100,
  no.~3, pp.~449--451, 2022.

\bibitem{shi2022multi}
W.~Shi, S.~Zhou, Z.~Niu, M.~Jiang, and L.~Geng, ``Multi-user co-inference with
  batch processing capable edge server,'' {\em IEEE Trans. Wireless Commun.},
  pp.~1--1, 2022.

\bibitem{hanhirova2018latency}
J.~Hanhirova, T.~K{\"a}m{\"a}r{\"a}inen, S.~Sepp{\"a}l{\"a}, M.~Siekkinen,
  V.~Hirvisalo, and A.~Yl{\"a}-J{\"a}{\"a}ski, ``Latency and throughput
  characterization of convolutional neural networks for mobile computer
  vision,'' in {\em Proceedings of the 9th ACM Multimedia Systems Conference},
  pp.~204--215, 2018.

\bibitem{szegedy2015going}
C.~Szegedy, W.~Liu, Y.~Jia, P.~Sermanet, S.~Reed, D.~Anguelov, D.~Erhan,
  V.~Vanhoucke, and A.~Rabinovich, ``Going deeper with convolutions,'' in {\em
  Proceedings of the IEEE conference on computer vision and pattern
  recognition}, pp.~1--9, 2015.

\bibitem{little1961proof}
J.~D. Little, ``A proof for the queuing formula: L = $\lambda$w,'' {\em
  Operations Research}, vol.~9, no.~3, pp.~383--387, 1961.

\bibitem{puterman2014markov}
M.~L. Puterman, {\em Markov decision processes: discrete stochastic dynamic
  programming}.
\newblock John Wiley \& Sons, 2014.

\bibitem{sennott1989average}
L.~I. Sennott, ``Average cost {semi-Markov} decision processes and the control
  of queueing systems,'' {\em Probability in the Engineering and Informational
  Sciences}, vol.~3, no.~2, pp.~247--272, 1989.

\bibitem{thomas1985finite}
L.~Thomas and D.~Stengos, ``Finite state approximation algorithms for average
  cost denumerable state {Markov} decision processes,'' {\em
  Operations-Research-Spektrum}, vol.~7, no.~1, pp.~27--37, 1985.

\end{thebibliography}
